\documentclass{article}

\usepackage[final]{corl_2020} % Uncomment for the camera-ready ``final'' version.
\usepackage{graphicx}
\usepackage{comment}
\usepackage{amsmath,amssymb} % define this before the line numbering.
\usepackage{color}
\usepackage{booktabs}
\usepackage[dvipsnames]{xcolor}
\usepackage{multirow}
\usepackage{wrapfig}

\title{Recovering and Simulating Pedestrians in the Wild}

\author{
  Ze Yang$^{1,2}$, Siva Manivasagam$^{1,2}$, Ming Liang$^{1}$, Bin Yang$^{1,2}$, Wei-Chiu Ma$^{1,3}$, Raquel Urtasun$^{1,2}$\\
  Uber Advanced Technologies Group$^{1}$,~ University of Toronto$^{2}$,~ MIT$^{3}$\\
  \texttt{\{zey,manivasagam,ming.liang,byang10,weichiu,urtasun\}@uber.com}
}

\newcommand{\bD}{\mathbf{D}}

\begin{document}
\maketitle

%===============================================================================

\begin{abstract}
	% !TEX root =  ../main.tex
Sensor simulation is a key component for testing the performance of self-driving vehicles 
and for data augmentation to better train  perception systems. 
Typical approaches rely on artists to create both 3D assets and their animations to generate a new scenario. 
This, however, does not scale. 
In contrast, we propose to recover the shape and motion of pedestrians from sensor readings captured in the wild by a self-driving car driving around.  
Towards this goal, we formulate the problem as energy minimization in a deep structured model that exploits human shape priors, reprojection consistency with 2D poses extracted from images, and a ray-caster that encourages the reconstructed mesh to agree with the LiDAR readings. 
Importantly, we do not require any ground-truth 3D scans or 3D pose annotations. 
We then incorporate the reconstructed pedestrian assets bank in a realistic LiDAR simulation system by performing motion retargeting, and show that the simulated LiDAR data can be used to significantly reduce the amount of annotated real-world data required for visual perception tasks.

\end{abstract}

% Two or three meaningful keywords should be added here

%===============================================================================

\vspace{-10pt}
% %!TEX root = ../main.tex
\section{Introduction}
\vspace{-5pt}

A key requirement for mobile robots is that they interact and maneuver safely around humans.
This is especially the case in autonomous driving, where the self-driving car should perceive in 3D each pedestrian in the scene and forecast their future trajectories.
To deploy in the real world, we must verify that our autonomy system is robust and handles safety-critical cases such as a child occluded by a bus running in front of the car.
However, it is unethical to test such cases in the real-world.
Moreover, it is expensive and not scalable to collect and manually label the full distribution of pedestrian scenarios to generate training and testing data for current ML-based perception systems.

An appealing alternative is leveraging realistic sensor simulation systems to train and test the perception system. Here we focus on simulating realistic traffic scenes with pedestrians for the LiDAR sensor, a common sensor in self-driving.
However, pedestrians are especially difficult to simulate; unlike vehicles, they are non-rigid objects that have a wide variety of shape, poses, and behaviors.

There are two lines of work when it comes to sensor simulation of pedestrian assets.
One approach is to use artist-designed human meshes (e.g., CARLA \cite{CARLA}).
Another is to use high-end 3D scanning systems in a controlled lighting setting with multiple cameras and/or depth sensors to create high-resolution human meshes \cite{h36m_pami, bogo2014faust, sigal2010humaneva, trumble2017total}.
Both approaches require an artist to ``rig" and animate behaviors for each human, which requires significant effort: the artist must first add a skeleton to the mesh for skinning and posing the character and then design the sequence of joint angles for the pedestrian skeleton required to simulate a particular behavior.
While these approaches have been widely used for creating realistic looking pedestrians in video games and movies, they are expensive and not scalable: it is difficult to manually create or 3D scan all the diverse variations in shape, pose, and trajectories a pedestrian may take in the real-world.

There has also been a large body of prior work on estimating 3D pose and shape from single images \cite{SMPLify, balan2007detailed, HMR, kolotouros2019learning, alldieck2019tex2shape} or video \cite{arnab2019exploiting, humanMotionKanazawa19, alldieck2018video, kocabas2020vibe}.
This is a more scalable solution, as images and videos of people are everywhere.
However, image-only  methods are prone to having incorrect location/movement estimates in 3D and can sometimes produce unrealistic looking meshes due to inaccurate depth estimates.
As a consequence, while they have produced visually appealing results, which might be sufficient in some application domains (e.g.,  augmented reality,  online  games), their 3D fidelity is not sufficient when simulating pedestrian LiDAR readings.

\begin{figure}[t]
\vspace{-25pt}
    \centering
    \includegraphics[width = 0.92\linewidth]{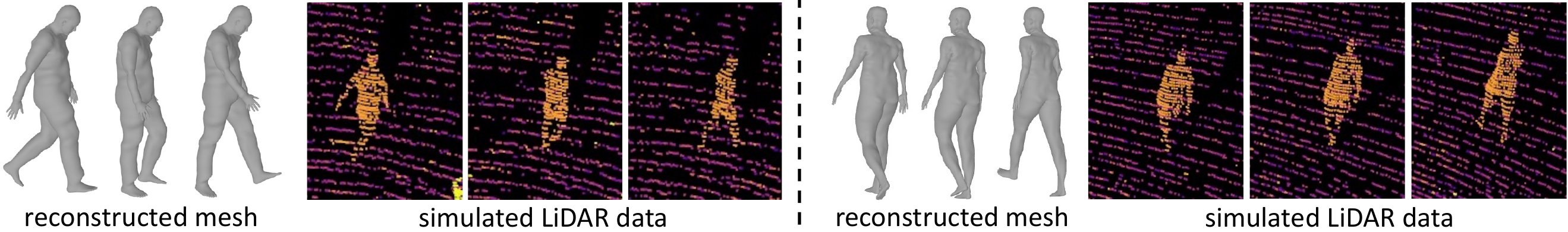}
    \vspace{-8pt}
    \caption{We recover realistic 3D human meshes and poses from sequences of LiDAR and camera  readings, which can then be used in sensor simulation for perception algorithm training and testing.}
	\vspace{-10pt}
    \label{fig:demo}
\end{figure}

Towards this goal, we leverage real-world sensor data captured by our autonomous driving fleet, which contain LiDAR point clouds and camera image sequences, to recover accurate 3D motion and shapes of pedestrians.
Our approach, \textbf{Li}DAR for human \textbf{M}esh \textbf{E}stimation (LiME), only requires a single low-cost artist-created mesh that we exploit to create a prior over human shapes, which we then pose and deform to match the sensor data.
We leverage the power of both deep learning and energy minimization methods to accurately recover shape and pose in the wild when no ground-truth is available.
To simulate the virtual world, we use a realistic LiDAR simulation system, LiDARsim \cite{siva2020lidarsim}, which uses real world data to generate a large collection of realistic background meshes of different scenes as well as vehicle assets. We then enhance it with a diverse bank of pedestrian shapes and poses reconstructed in the wild using LiME.
We can then generate novel scenarios by selecting pedestrians in our bank, applying motion retargeting, and placing them in the scene.
LiDARsim then renders the scene to generate realistic LiDAR point clouds.
We show that we can generate simulated LiDAR that has little to no sim2real domain gap, which allows us to evaluate  a state-of-the-art perception system.
Furthermore, we demonstrate that when generating low-cost simulation data at scale for training data augmentation, we reduce the need for labeled data.

\vspace{-5pt}
\section{Related Work}
\vspace{-0.2cm}
\paragraph{3D Human Pose and Motion Estimation:}
Human motion capture (MoCap) is usually conducted in highly-calibrated and laboratory-controlled environments. With the help of multi-view sensing \cite{joo2018total} and marker-based technology \cite{h36m_pami}, many high-quality dynamics measurements \cite{sigal2010humaneva, trumble2017total, CMU_mocap} have been collected, including accurate 2D and 3D skeletal joint locations over time.
Based on these datasets, several methods have been developed to predict 3D human pose and motion from monocular images \cite{agarwal2005recovering, Martinez_2017_ICCV, pavlakos2018ordinal} and monocular video \cite{sigal2012loose, pavllo20193d, tekin2016direct}, achieving state-of-the-art performance.
Unfortunately, these data, while useful, are still over-simplified. Numerous real world scenarios are not captured, $\emph{e.g.},$ environmental occlusions.
To overcome such limitations, recent work has focused on capturing large scale ``in-the-wild'' datasets with 3D pose using IMU and cameras \cite{von2018recovering, saini2019markerless, vonMarcard2018}.
Most efforts still focus on pose estimation from images. However, they have difficulty obtaining precise shape and pose because accurate depth is missing.
We require more accuracy for simulating pedestrian scenarios and testing autonomy.
Recent work \cite{bogo2015detailed,zimmermann20183d} have proposed using RGB-D images to predict 3D pose in indoor environments, but to our knowledge, we are the first to tackle estimating 3D pose over time from images and sparse LiDAR points at distance.
This setting is important for recovering and simulating realistic humans in the wild for self-driving.

\vspace{-5pt}
\paragraph{Non-rigid Body Surface Reconstruction:}
For realistic simulation, we need to reconstruct the 3D human mesh in the scene.
While real-time mesh reconstruction of non-rigid objects from depth camera \cite{DynamicFusion} or RGB camera \cite{PIFU,DeepHuman} exist, to re-articulate the humans for downstream tasks we also require human pose.
We now discuss past work that recover both 3D pose and articulated meshes.
Most of these work \cite{SMPLify, balan2007detailed, HMR, kolotouros2019learning, alldieck2019tex2shape} rely on strong shape priors such as SMPL~\cite{SMPL}.
They either directly regress human model parameters from observations or fit the parametric model to RGB images by minimizing carefully designed energy functions. To further ensure temporal consistency, \cite{arnab2019exploiting, humanMotionKanazawa19, alldieck2018video, kocabas2020vibe} leverage training signals from videos.
\cite{li2019lbs} align articulated models with free-form deformation on densely sampled point clouds from multiple sensors.
We focus on extending the recovery of 3D human pose and shape  with small error from partial LiDAR data and images.

\vspace{-5pt}
\paragraph{Sensor simulation of Pedestrians:}
Prior work \cite{CARLA}, simulate pedestrians by first creating artist-designed meshes that are manually rigged and animated, and then performing sensor simulation via graphics engine rendering. 
While this allows for fine-grained control of pedestrian appearance and behavior, it is a time-consuming and expensive, which does not scale to capture the real-world pedestrian distribution.
Efforts have been made to incorporate human avatars into robot simulators such as MORSE \cite{echeverria2011modular} for prototyping, data collection and evaluation \cite{lemaignan2014simulation}. 
This has focused mostly on indoor-enviroments and with unrealistic rendering. 
Our work focuses on leveraging data from in-the-wild, where we can automatically capture diverse human appearance, motion, and behavior and directly adapt these assets for realistic sensor simulation.

\vspace{-5pt}
% !TEX root =  ../main.tex
\section{Human Model}
\label{sec:human_shape}
\vspace{-0.3cm}

We utilize a Linear Blend Skinning (LBS) model, which we enhanced with both bone scaling and per-vertex deformations to represent how the human body deforms as a function of its pose.
We use this enhanced LBS model due to its simplicity and efficient computation, as opposed to higher-order blend skinning methods such as spherical \cite{kavan2005spherical} or non-skeleton based deformation methods~\cite{joshi2007harmonic}.
Our experiments show that this simple representation outperforms popular human models (e.g., SMPL \cite{SMPL}) in reconstructing 3D shape from sensor data.  Furthermore, it proves sufficient for our downstream task of simulating LiDAR data for testing and improving perception algorithms.
We now review the LBS model and describe our bone-scaling and per-vertex deformation modifications to handle shape variation and appearance.

LBS represents the human body in two parts: a mesh representation and a hierarchical set of interconnected bones, {\em i.e.}, the skeleton.
The key idea is that as the skeleton moves, the positions of the mesh's vertices will change, but not their connectivity.
Each bone in the skeleton is associated with some portion of the character's visual representation (i.e., set of vertices) in a process called {\it skinning}.
Each mesh vertex has a specific corresponding ``blend weight" for each skeleton bone.
To calculate the final position of a vertex, a transformation matrix is created for each bone which, when applied to the vertex, first puts the vertex in bone space, and then puts it back into mesh space.
After applying this transformation to the vertex, it is scaled by its corresponding blend weight.

More formally, let the template mesh  $\mathbf{V}\in \mathbb{R}^{N\times 3}$ be the set of $N$ vertices $\mathbf{V} =\{\mathbf v_i \}_{i=1}^N$  (with oriented normals $\mathcal N =\{\mathbf n_i \}_{i=1}^N $), and let $\mathbf{W}\in \mathbb{R}^{N\times K}$ be the set of blend weights\footnote{These blend weights can be created for example by diffusing artist-annotated part-segmentations \cite{SMPL}.}.
We represent a skeleton pose with the set of joint rotation matrices $\mathbf \Theta_i \in \textbf{SO}(3)$, one for each joint representing the rotation with respect to its parent in the skeletal tree.
While this original LBS formulation is a good approximation of the human skeleton, it cannot model well different human body sizes  deviating from the template mesh.
To address this, we introduce a learnable scale factor for each bone in the skeleton: where $s_p$ denotes the bone length scale factor between the $p$-th joint and its parent, which we model to be symmetric with respect to the human spine, {\em e.g.}, the left and right arms share the same bone scale factor.
We thus traverse the tree and construct the transformation matrix for each joint $\mathbf T_k(\mathbf\Theta) \in \textbf{SE}(3)$:
\vspace{-5pt}
\begin{align}
    \label{eq::chain_scale}
    \mathbf T_k(\mathbf{s}, \mathbf\Theta) &= \prod_{p \in A(k)}
    \begin{bmatrix}
    {s_p} \mathbf \Theta_p & \mathbf (\mathbf I - s_p \mathbf \Theta_p) \mathbf j_p \\
    \mathbf 0 & 1
    \end{bmatrix}
\end{align}
where $A(k)$ is the set of  joint ancestors of the $k$-th joint in order, $\mathbf \Theta_p$ is the rotation matrix of the $p$-th joint wrt its parent, and $\mathbf j_p$ is the coordinate of the $p$-th joint in the template mesh.
The coordinate for the $i$-th vertex can now be computed as a linear combination of the joint transformation matrices and its unique blend weights.  However, the template mesh vertices alone cannot handle shape variations. Therefore, following \cite{li2019lbs}, we also add a displacement vector for each vertex. The coordinate for the $i$-th vertex and the $k$-th joint in the posed mesh are computed as:
\begin{align}
    \mathbf{\bar{v}}_i=\sum_{k=1}^K \mathbf T_k (\mathbf{s}, \mathbf \Theta) (\mathbf v_i + \mathbf n_i {d_i}) ~w_{i,k} + \mathbf c \,\,,
    \quad \quad
    \mathbf{\bar{j}}_k = \mathbf T_k(\mathbf{s}, \mathbf\Theta) \mathbf{j}_k + \mathbf c
    \label{eq::lbsj}
\end{align}
where $w_{i,j}$ is the skinning weight describing the influence of the $k$-th joint on the $i$-th vertex in the template shape, and $\mathbf c \in \mathbb R^3$ is the global translation of the root joint. 
The final posed mesh model is
\begin{equation}
\mathbf M = \mathcal M(\mathbf W, \mathbf V, \mathcal N, \mathbf \Theta, \mathbf c, \mathbf s,  \mathbf D)
\label{eqn::human_model}
\end{equation}
with posed mesh $\mathbf M$, blend weights $\mathbf W$,  mesh vertices $\mathbf V$, normals $\mathcal N$,  joint angles $\mathbf \Theta$, root location $\mathbf c$, bone scale factors $\mathbf s$, and per-vertex deformation matrix $\mathbf D$.

\vspace{-5pt}
\section{Reconstructing Pedestrians in the Wild}
\label{sec::reconstruction}
\vspace{-0.3cm}

\begin{figure}[t]
\vspace{-0.7cm}
    \centering
    \includegraphics[width = 1.0\linewidth]{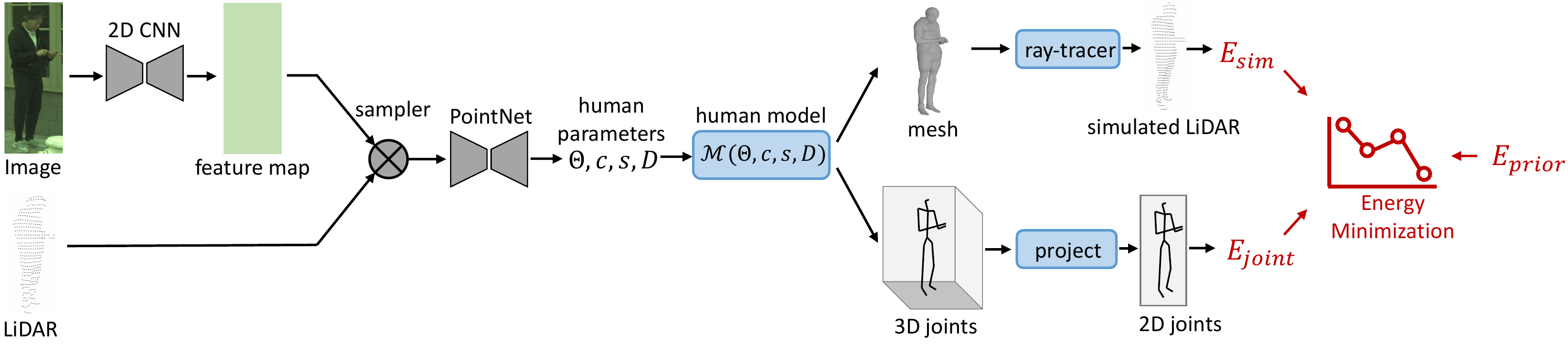}
    \vspace{-0.55cm}
    \caption{
    LiDAR for human Mesh Estimation, (LiME): Given sensory observations, a sensor fusion regression network predicts the human parameters which minimize the objective function in Eq.~\ref{eq::energy_minimization}. 
    We then perform energy minimization over the sequence to obtain an optimized shape and 3D pose.
	}
	\vspace{-10pt}
    \label{fig:architecture}
\end{figure}

We now describe our method, \textbf{Li}DAR for human \textbf{M}esh \textbf{E}stimation (LiME), for reconstructing pedestrians in the wild.
Given a sequence of LiDAR measurements and camera images captured by a self-driving car, as well as 3D bounding boxes enclosing the pedestrians we want to reconstruct, we seek to estimate the pose trajectory (including global motion) and shape of each pedestrian in the scene.
We use our modified LBS model $\mathcal M$ defined in Eq.~\ref{eqn::human_model} as our human body parameterization.
For our reconstructions, the body model's skinning weights $\mathbf{W}$,  template shape $\mathbf{V}$ and  normals $\mathcal{N}$ are fixed and we infer from data the pose (joint angles $\mathbf \Theta$,  offset $\mathbf c$) and shape modifications (joint scale factors $\mathbf s$ and deformations $\bD$).
We first use a regression network to predict the initial estimates of ($\mathbf \Theta, \mathbf c, \mathbf s,  \mathbf D$) from data.
We then perform energy minimization to refine the prediction (see Figure~\ref{fig:architecture}).
As we do not have ground-truth pose or shape, we use the objective function to self-supervise our network.
We now describe the regression network and energy minimization in more detail.

\vspace{-0.3cm}
\subsection{Sensor Fusion Regression Network}
\label{sec:sfrn}
\vspace{-0.3cm}
Our regression network takes as input the LiDAR and camera image centered and cropped around each pedestrian, and outputs the initial estimate of the  body model parameters ($\mathbf \Theta, \mathbf c, \mathbf s,  \mathbf D$).
Towards this goal, the camera image is fed into a 2D CNN to compute image features.
We then apply bilinear interpolation to sample the corresponding image feature for each LiDAR point using geometry and the camera calibration.
Finally, each LiDAR point and its concatenated image feature are consumed by a PointNet~\cite{PCN} network to predict the human parameters.
Since the regression network has difficulty identifying which direction the human is facing, we follow \cite{insafutdinov2018unsupervised} and run two branches of the network, where the root joint angle is initialized to either face forward (0$^{\circ}$) or backward (180$^{\circ}$). 

\vspace{-0.3cm}
\subsection{Energy  Formulation}
\vspace{-0.3cm}
We define the objective function to capture the fact that our shape should be consistent with the point clouds from the LiDAR measurements ($E_{\text{sim}}$) and that the estimated 3D joints should be consistent with the 2D joints estimated from images ($E_{\text{joint}}$).
We add an additional term, $E_{\text{prior}}$, to regularize the poses to be natural, and the deformed shape to be smooth and not have large deviations from the mesh template. The full objective function is:
\begin{align}
E(\mathbf \Theta_{1:T}, \mathbf c_{1:T}, \mathbf s,  \mathbf D)=
 \sum_t \lambda_{\text{sim}}  E_{\text{sim}}(\mathbf \Theta_t, \mathbf c_t, \mathbf s,  \mathbf D)
+ \lambda_{\text{joint}} E_{\text{joint}}(\mathbf \Theta_t, \mathbf c_t, \mathbf s)
+ E_{\text{prior}}(\mathbf \Theta_t, \mathbf s,\mathbf D)
\label{eq::energy_minimization}
\end{align}
where $t$ is the time step in the pedestrian trajectory, and $\mathbf \Theta_{1:T}$, $\mathbf c_{1:T}$ are the sequence of pose joint angles and root offsets.
We next describe how we compute each term. 

\vspace{-0.3cm}
\paragraph{LiDAR Consistency:} 
The LiDAR consistency term encourages the ray-casted point cloud from the estimated mesh $M(\mathbf \Theta_t, \mathbf c_t, \mathbf s,  \mathbf D)$ to match with the real partial point cloud $\mathbf X$ of the pedestrian through the Chamfer loss:
\begin{align}
    \label{eq::sim}
    E_{\text{sim}}(\mathbf \Theta_t, \mathbf c_t, \mathbf s,  \mathbf D) &= \frac{1}{|\mathbf X|} \sum_{\mathbf x \in \mathbf X} \min_{\mathbf y \in \mathbf Y} \left\lVert \mathbf x - \mathbf y \right\rVert_2^2 +
	\frac{1}{|\mathbf Y|} \sum_{\mathbf y \in \mathbf Y} \min_{\mathbf x \in \mathbf X} \left\lVert \mathbf y - \mathbf x \right\rVert_2^2
\end{align}
where $|\mathbf X|$ denotes the cardinality of point set $\mathbf X$, and $\mathbf Y = \{y_1 \dots y_n | y_i \in \mathbb R^3\}$ is the rendered points from the estimated mesh.
Note that this is a differentiable point set distance and  we exploit the Moller-Trumbore~\cite{RayCasting} ray casting algorithm which is differentiable (w.r.t. the mesh vertices)  such that the full model can be trained end-to-end. We refer the reader to the supplementary material for details of the ray-caster and its differentiability.
When computing $E_{\text{sim}}$, we take into account objects that occlude the sensor's field-of-view of the pedestrian, thereby ignoring simulated points from the ray-caster that would not appear due to occlusion.

\vspace{-0.3cm}
\paragraph{Human Joints Consistency:}
We exploit camera images by first detecting 2D joints using a state-of-the-art 2D pose estimator \cite{wu2019detectron2}. We then encourage the projection of the predicted 3D pose to be consistent with the 2D pose estimates:
\begin{align}
\label{eq::joints}
E_{\text{joint}}(\mathbf \Theta_t, \mathbf c_t, \mathbf s) = \sum_{k \in B} m_k \rho (\pi(\mathbf j_k, \mathbf \Omega) - p_k)
\end{align}
where $\mathbf j_k$ is the $k$-th joint transformed according to Eq. \ref{eq::lbsj}, $B$ is the subset of 3D joints that have 2D counterparts, and $p_k$ and $m_k$ are the corresponding estimated 2D joint and confidence score.
$\pi$ is the projection function that takes the  camera parameters  $\mathbf \Omega$, which are given as cameras of self-driving cars are calibrated,  and projects the 3D joint locations onto the image plane.
$\rho$ is the $\sigma^2$-scaled Geman-McClure robust penalty function  defined as $\rho(x) = (x^2 * \sigma ^2) / (x^2 + \sigma^2$), with $\sigma=100$.

\vspace{-0.3cm}
\paragraph{Pose and Shape Priors:}
We incorporate our prior knowledge of what are reasonable human poses and shapes to be robust to noisy sensor data.
For joint angles, we follow \cite{SMPLify, arnab2019exploiting} and represent the joint angle prior as the negative log-likelihood of a Gaussian Mixture Model (GMM) learned from the CMU Mocap dataset~\cite{CMU_mocap}.
We also add a bone scale prior that encourages the bone length to be close to a canonical size.
The pose prior is:
\begin{align}
    \label{eq::pose_prior}
    E_{\text{pose}}(\mathbf \Theta_t, \mathbf s) = -(\log(\sum_r^R g_r \mathcal N(\mathbf \Theta; \mathbf \mu_r, \mathbf \Sigma_r))) + \lambda \sum_k^K ( \prod_{p\in A(k)}s_p - 1)^2
\end{align}
with $R=8$ Gaussians, $(g_r, \mathbf \mu_r, \mathbf \Sigma_r)$  the weight, mean and covariance of the $p$-th Gaussian, and $ \prod_{p\in A(k)}s_p$  the cumulated scale factor for the bone length between the $k$-th joint and its ancestors.
To ensure the deformed mesh still retains most of the mesh template shape and has smoothly-varying and small deformations, we add a Laplacian mesh regularizer~\cite{sorkine2004laplacian} and $\ell_2$ regularizer, respectively:
\begin{align}
    \label{eq::shape_prior}
    E_{\text{shape}}(\mathbf D) = \sum_{i=1}^N \lVert \mathcal L(\mathbf v_i + \mathbf n_i d_i) - \mathcal L(\mathbf  v_i) \rVert_2^2 +\lambda \sum_{i=1}^N d_i^2
\end{align}
where $\mathbf v_i$ and $\mathbf n_i$ are the vertex location and normal in the mesh template, $d_i$ is the corresponding displacement along the normal direction, and  $\mathcal L$ is the Laplace operator.
The total prior is:
\begin{align}
    \label{eq::prior}
    E_{\text{prior}}(\mathbf \Theta_t, \mathbf s,\mathbf D) = \lambda_{\text{pose}} E_{\text{pose}}(\mathbf \Theta, \mathbf s) + \lambda_{\text{shape}} E_{\text{shape}}(\mathbf D)
\end{align}

\vspace{-0.4cm}
\subsection{Learning and Inference}
\vspace{-5pt}

\paragraph{Inference:} 
We perform a forward pass for each pedestrian frame and output the initial model parameters. These predictions are then further refined by minimizing  the differentiable  energy defined in Eq. \ref{eq::energy_minimization}, which obtains the final pose and shape of each pedestrian at each frame.
In practice we found that a two step energy minimization works well, where we first optimize $\mathbf \Theta_{1:T}$, $\mathbf c_{1:T}$, $\mathbf s$ till convergence, and then optimize the deformation variable $\mathbf D$ till convergence.
Each substep converges in typically 50 iterations.
We adopt the Adam optimizer \cite{kingma2014adam}, which ran much faster than a second-order optimizer, to optimize our objective.
Please see supplementary for more details.

\vspace{-5pt}
\paragraph{Learning:}
Since we do not have ground-truth shape or pose for our in the wild setting, we use Eq.~\ref{eq::energy_minimization} (for a single frame) as the loss function to train the network in a self-supervised fashion. 
As mentioned in Section~\ref{sec:sfrn} we use two branches with different root initializations, and perform hindsight loss during training \cite{insafutdinov2018unsupervised}. 
We pass the result with the lower loss to the energy minimization step during inference.
Please see supplementary for more details.

\vspace{-5pt}
% !TEX root =  ../main.tex
\section{LiDAR Simulation of Pedestrians}
\begin{figure}
\vspace{-0.7cm}
\centering
\begin{minipage}{.50\textwidth}
  \centering
 \includegraphics[width=0.9\textwidth]{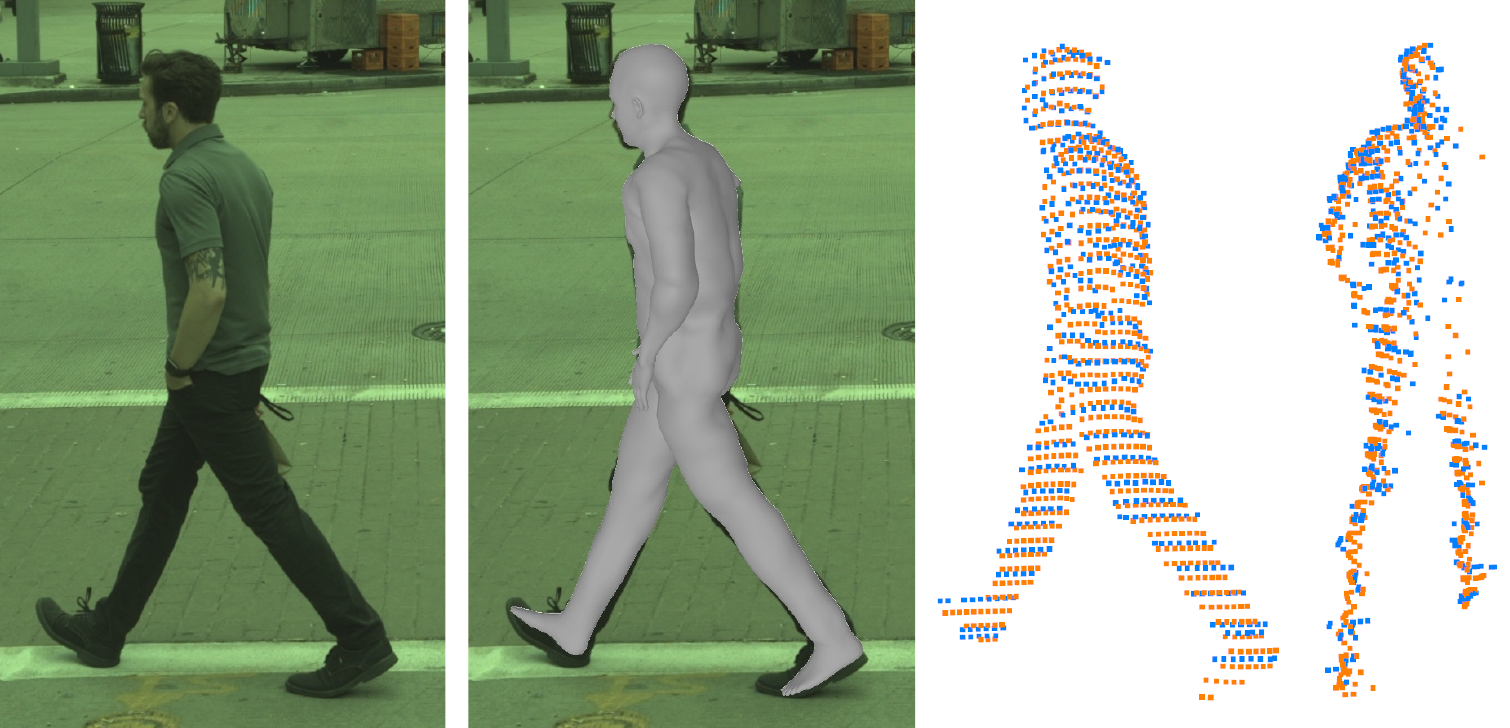}
  \caption{Result on real world data using LiME:
(1) Camera image. (2) Reconstructed mesh. (3) \textcolor{orange}{Ray-casted points} on recovered mesh, overlapped with \textcolor{blue}{GT LiDAR points}. (4) Side view.}
\label{fig::vis_overlap}
\end{minipage}
\hfill
\begin{minipage}{.47\textwidth}
  \centering
    \includegraphics[width=\textwidth]{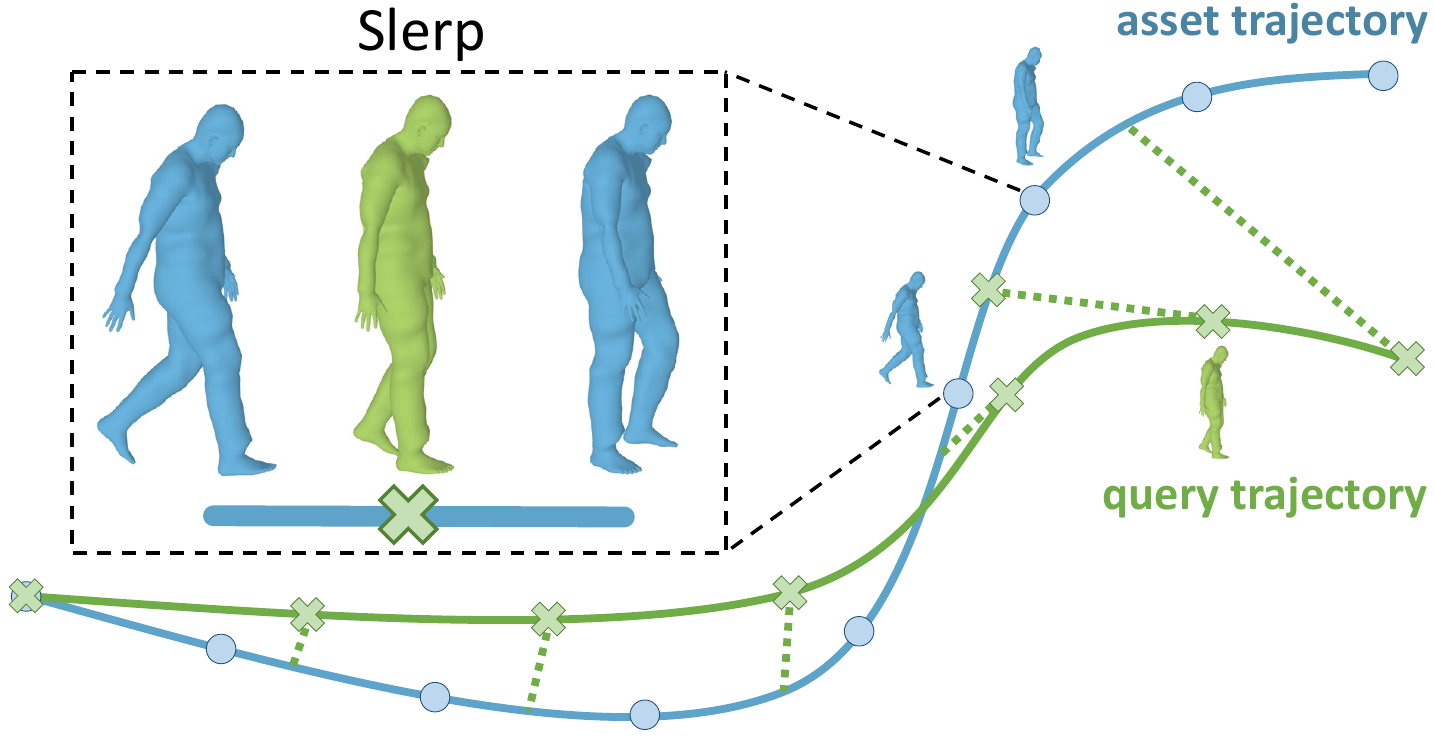}
  \caption{Interpolating a mesh for a query trajectory from recovered asset mesh sequences.}
    \label{fig::slerp}
\end{minipage}
\vspace{-15pt}
\end{figure}

\label{sec::simulation}
\vspace{-0.2cm}

In order to produce realistic sensor simulation, we first require a virtual world with realistic backgrounds (i.e. roads, buildings, signs) and dynamic objects (e.g., vehicles, pedestrians), as well as a sensor simulation system that has high-fidelity with the real LiDAR. 
LiDARsim \cite{siva2020lidarsim} is a LiDAR simulator that uses real data to generate background scenes and vehicle dynamic objects. 
LiDARsim then places the assets in a scenario (provided by either a labeled snippet, a tracking algorithm, artist-drawn trajectories, or algorithmically) and renders realistic LiDAR at each time step using both physics and machine learning. In particular, a neural network is used to enhance the realism of the ray-casted LiDAR by determining which points would not return in the real world (e.g., due to spectral reflections or far distance).
We note that this simulator is different from the ray-tracer desribed in Section~\ref{sec::reconstruction}, as LiDARsim has a high-performing ray-tracer to scale to  millions of scene elements.  This raytracer is also non-differentiable, thus not suited for our reconstruction framework. 
While LiDARsim provides high-fidelity backgrounds and vehicles, it lacks realistic pedestrians.

We now describe how we enhance LiDARsim with the pedestrians reconstructed using LiME.
Towards this goal, we first build an asset bank of pedestrian sequences and their corresponding meshes directly from data captured by our self-driving fleet.
Since the trajectories and mesh sequences in the asset bank can be quite diverse (walking, running, standing, sitting, etc.), we ease the reuse of the  action-specific pose dynamics by clipping each cyclic pedestrian trajectory to consist of a single cycle, where the human poses in the start and end frames are similar.
The average action cycle length is 1.5 seconds.
Then, for each new query pedestrian trajectory to be simulated, we select a pedestrian moving at a similar speed from the asset bank, adapt it to the new scene, and simulate the LiDAR data with LiDARsim.
We now discuss each step in more detail.

Our simulation approach works as follows: 
The user provides a bird's eye view (BEV) 2D trajectory in the scene map as a high level description of the motion to simulate.
Note that this trajectory can come from an existing trajectory (recovered from 
recorded snippets via tracking or labeling), can be drawn by a test engineer, or can be produced algorithmically. 
In our experiments, we use labeled snippet trajectories as our query trajectories.
We then retrieve the asset in the bank which is most similar to this trajectory query. 
We use velocity as our similarity function (specifically, the asset trajectory whose velocity is consistently within 0.5 m/s of the query trajectory's), as action-specific pose dynamics are specific to particular velocities. 
We then modify the retrieved asset and retarget it to perform the desired motion.
Specifically, we project the query trajectory to the retrieved asset trajectory in BEV, and use Slerp \cite{shoemake1985animating} to interpolate the human poses for each time-step in the query trajectory. Note that this modification affects both the joint angles and associated mesh via our skinning model (see Fig.~\ref{fig::slerp}).
Finally, we use LiDARsim to simulate the scene as seen by the sensor.

\vspace{-5pt}
% %!TEX root = ../main.tex
\section{Experimental Evaluation}
\label{sec::experiments_recon}
\vspace{-10pt}
We first evaluate our proposed method for estimating human surface geometry from LiDAR and image sequences.
We then show how capturing realistic pedestrian trajectories in the wild enhances simulation environments and improves performance on autonomy tasks such as pedestrian detection.

\vspace{-0.3cm}
\subsection{Pedestrian Reconstruction from Sparse LiDAR Points}
\vspace{-0.3cm}
We evaluate our model on the 3DPW \cite{vonMarcard2018} dataset, which contains 60 sequences (12 validation split) of real world scenarios and 18 different humans, with images and ground-truth pose and complete clothed 3D shape.
We place a virtual LiDAR sensor at the camera center and ray-cast the clothed human mesh in the dataset to generate simulated LiDAR points.
Given 3DPW real images and synthetic LiDAR, we evaluate our algorithm on estimating pose and shape.
We measure the mesh error in cm with the mean Per-Vertex-Error (PVE) and square root of the Chamfer distance (CD) between the vertices of our prediction vs. the ground-truth's.
We measure the joint estimation error with the mean Per-Joint-Position-Error (MPJPE) in cm.

\vspace{-5pt}
\paragraph{Ablation on input feature and energy minimization:}
The effect of using different input features is reported in Table~\ref{tab::input_feature}. 
When we fuse image and LiDAR, the reconstruction error is lower than using either feature. 
Energy minimization further reduces the error. 

\begin{table}[t]
\small
\begin{minipage}{0.51\textwidth}
    \begin{center}
    \begin{tabular}{cccc}
\toprule[0.1em]
    %\hline
    & ~CD~ & PVE & MPJPE \\
    %\hline
    \midrule
    Image & 6.77 & 14.26 & 12.16 \\
    %\hline
    LiDAR & 4.94 & 11.17 & 9.51\\
    %\hline
    Fused 
& 4.37 & 9.30 & 7.98 \\
    %\hline
    \midrule
    Fused + EM & \textbf{2.17} & \textbf{5.78} & \textbf{5.01} \\
\bottomrule[0.1em]
    \end{tabular}
    \end{center}
    \caption{Effect of input/energy minimization (EM)}%The results are reported with $cm$.}
    \label{tab::input_feature}
\end{minipage}
\hfill
\begin{minipage}{0.48\textwidth}
    \begin{center}
    \begin{tabular}{cccc}
\toprule[0.1em]
    Human Model & ~CD ~ & PVE & MPJPE \\
\midrule
    %\hline
    %\hline
    LBS & 2.62 & 6.49 & 5.69 \\
    %\hline
    SMPL & 2.44 & 6.04 & 5.17 \\
    %\hline
    LBS + bone scale & 2.38 & 5.97 & 5.19 \\
    %\hline
    Ours & \textbf{2.17} & \textbf{5.78} & \textbf{5.01} \\
    %\hline
\bottomrule[0.1em]
    \end{tabular}
    \end{center}
    \caption{Effect of different human model.}%The results are reported with $cm$.}
    \label{tab::human_model}
\end{minipage}
\vspace{-20pt}
\end{table}

\vspace{-5pt}
\paragraph{Alternate human model:} 
We study the effect of using different human models in Table~\ref{tab::human_model}. The results are reported with running energy minimization included. 
Using the LBS model alone, we achieve $6.49$ cm mean PVE.
With the additional bone scale factors and per-vertex displacement vectors, the PVE is $5.78$ cm, outperforming the SMPL model.

\vspace{-5pt}
\paragraph{Ablation on energy terms:}
Results in Table~\ref{tab::objective} are reported after running energy minimization. 
Leveraging LiDAR point cloud observations  is important to achieving lower Chamfer error.
Leveraging 2D joints is important to achieving lower mean PVE and MPJPE, which measure dense and sparse correspondence between our prediction and the ground-truth shape.
Each energy term contributes  to the final model.

\begin{figure}[t]
    \centering
    \includegraphics[width = 0.95\linewidth]{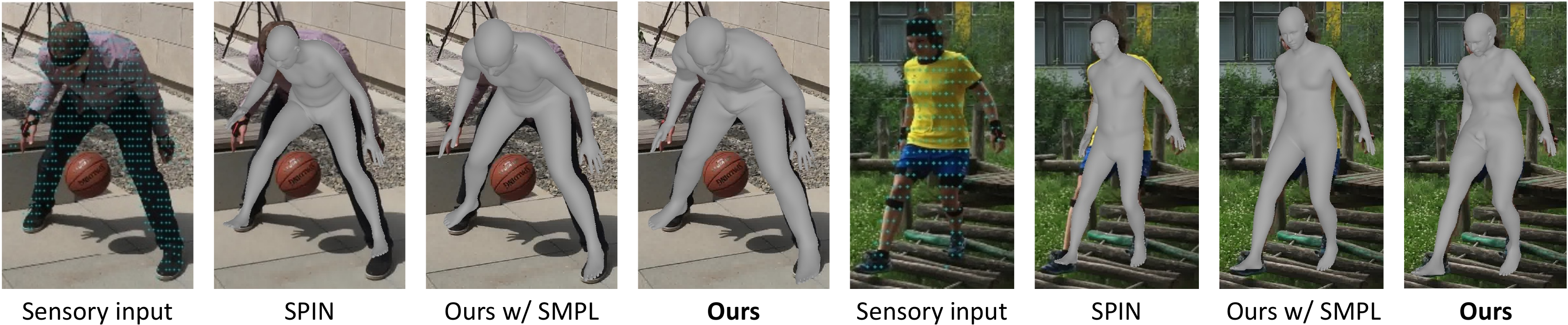}
    \vspace{-8pt}
    \caption{
    Quantitative results of our method on 3DPW~\cite{vonMarcard2018} dataset. The sensory input consists of camera image and the synthetic LiDAR points. We show our method using both SMPL model and our human model in Section~\ref{sec:human_shape}, and compare with SPIN~\cite{kolotouros2019learning}.
	}
	\vspace{-15pt}
    \label{fig:vis_recon}
\end{figure}

\vspace{-5pt}
\paragraph{State-of-the-art (SoTA) comparison:}
We compare our model with SoTA image-only approaches on the 3DPW~\cite{vonMarcard2018} test set in Table~\ref{tab::sota}.
``PVE*'' denotes the typically reported mean Per-Vertex-Error between prediction and ground-truth naked shape,  while ``PVE'' denotes the mean Per-Vertex-Error between prediction and ground-truth clothed shape, which is more relevant to our task.
We note that our approach uses sparse LiDAR, while other SoTA approaches uses ground-truth meshes and 3D poses and mix multiple datasets during training.
Figure~\ref{fig:vis_recon} shows qualitative results. Using LiDAR's sparse depth greatly improves the accuracy of the shape.

\begin{table}[t]
\small
\begin{minipage}{0.55\textwidth}
    \begin{center}
    \begin{tabular}{ccccccc}
\toprule[0.1em]
    \multicolumn{3}{c}{Objectives} & & \multicolumn{3}{c}{Error} \\
    %\cline{1-3}\cline{5-7}
\cmidrule{1-3}\cmidrule{5-7}
 %\midrule
    $E_{\text{sim}}$ & $E_{\text{prior}}$ & $E_{\text{joint}}$ && CD & PVE & MPJPE \\
\midrule
    %\hline
    %\hline
    \checkmark & & && 3.40 & 30.47 & 28.36 \\
    %\hline
    \checkmark & \checkmark && & 3.41 & 23.15 & 21.36 \\
    %\hline
    & \checkmark & \checkmark && 5.84 & 11.60 & 9.96 \\
    %\hline
    \checkmark & & \checkmark && 2.49 & 7.24 & 5.40 \\
    %\hline
    \checkmark & \checkmark & \checkmark && \textbf{2.17} & \textbf{5.78} & \textbf{5.01} \\
    %\hline
\bottomrule[0.1em]
    \end{tabular}
    \end{center}
    \caption{The ablation on different objective term for shape reconstruction and 3D joint estimation}
    \label{tab::objective}
\end{minipage}
\hfill
\begin{minipage}{0.44\textwidth}
    \begin{center}
    \begin{tabular}{cccc}
\toprule[0.1em]
    Methods & PVE* & PVE & MPJPE \\
\midrule
    %\hline
    %\hline
    HMMR~\cite{humanMotionKanazawa19} & 13.93 & --  & 11.65 \\
    %\hline
    SPIN~\cite{kolotouros2019learning} & 11.64 & --  & 9.69 \\
    %\hline
    VIBE~\cite{kocabas2020vibe} & 9.91 & --  & 8.29 \\
    %\hline
    Ours & \textbf{7.36} & \textbf{8.17} & \textbf{6.57} \\
    %\hline
\bottomrule[0.1em]
    \end{tabular}
    \end{center}
    \caption{Evaluation of 3D pose estimation and shape reconstruction on the 3DPW test set. ``PVE*" means Per-Vertex-Error when the ground-truth human is naked.}
    \label{tab::sota}
\end{minipage}
\vspace{-15pt}
\end{table}

% !TEX root =  ../main.tex
\vspace{-5pt}
\subsection{Simulation for Downstream Visual Application}
\vspace{-5pt}

We have demonstrated our approach on recovering human pose and shape from 3DPW pedestrian sequences in the wild, and that leveraging LiDAR point clouds in our energy formulation improves reconstruction performance over prior methods. 
We now leverage diverse and realistic pedestrians for downstream perception tasks.
We conduct our experiments on the ATG4D~\cite{yang2018pixor} self-driving dataset, which contains diverse scenes across multiple metropolitan cities in North America with bounding box labels annotated for each object in the scene. 
Each log snippet has 64-beam LiDAR sweeps at $10$ Hz for $\approx 25$ seconds with corresponding camera images.
We use a detector similar to PnPNet~\cite{pnpnet}, which takes as input five consecutive LiDAR sweeps in birds-eye-view (BEV) and outputs 3D bounding boxes for detected vehicles and pedestrians in the scene. More details about the detector can be found in the supplementary material.

We use LiME to reconstruct the pedestrian shape and pose trajectory from the ATG4D dataset.
LiME accurately captures the geometry compared to the original LiDAR sequence, as seen in Figure~\ref{fig::vis_overlap}.
To generate the assets in Section~\ref{sec::simulation}, we select 211 unique pedestrian trajectory annotations from the ATG4D~\cite{yang2018pixor} training split with over 3300 individually posed meshes.
Each selected pedestrian trajectory annotation has:
(1) visible camera images and $70\%$ of joint detection score $>$ 0.1; 
(2) has $\ge 10$ consecutive frames; 
(3) has $\ge 100$ number of LiDAR points per frame;
(4) $E_{sim} < 20$, $E_{joint} < 6$, and $E_{joint} < 22$;
(5) Forms a complete action cycle (Sec. \ref{sec::simulation}).

\begin{figure}[t]
    \centering
    \includegraphics[width = 1.0\linewidth]{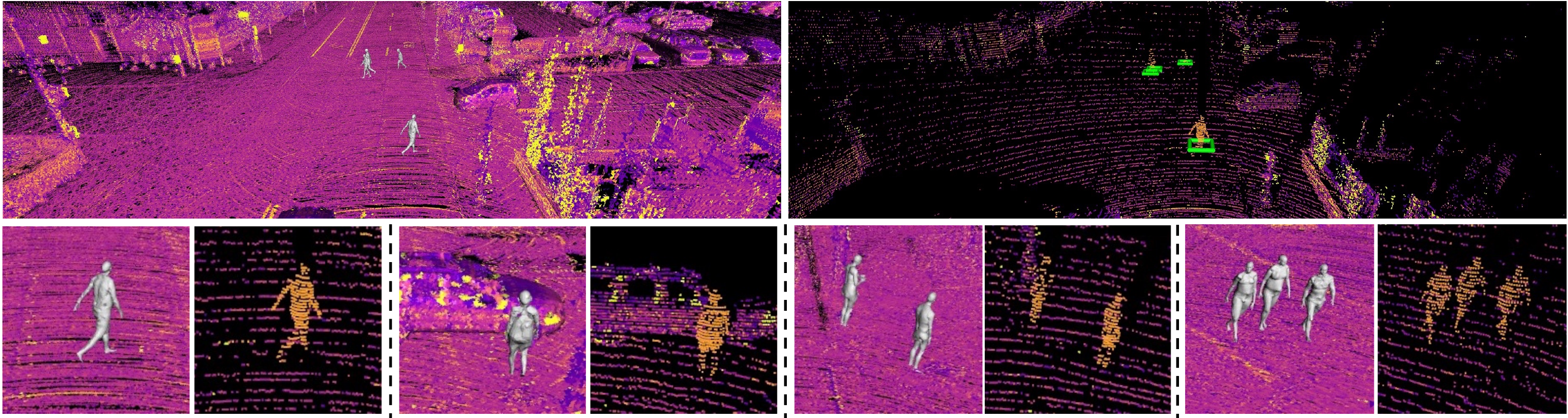}
    \vspace{-18pt}
    \caption{\textbf{Top Left:} reconstructed scene. \textbf{Top Right:} simulated LiDAR and pedestrian detections (green box). Detector trained on real-data only.
    \textbf{Bottom:} Reconstructions and simulated LiDAR.
}
    \vspace{-5pt}
    \label{fig::lidar_sim}
\end{figure}

We then use the method described in Section~\ref{sec::simulation} to generate simulated LiDAR sweeps,
as seen in Figure~\ref{fig::lidar_sim}. 
We can place pedestrians in new configurations (bottom panel one), generate occlusion (panel two), sample safety-critical behaviors (looking at phone, panel three), or create group behavior (panel four).
We show through data augmentation experiments that training on our simulated LiDAR data improves pedestrian detection performance with limited amounts of real data.

\begin{table}[t]
\small
\begin{minipage}{0.43\textwidth}
    \begin{center}
    \begin{tabular}{ccc}
\toprule[0.1em]
   % \hline
    Eval data & $\text{AP}_{0.3}$ & $\text{AP}_{0.5}$ \\
\midrule    
%\hline
    %\hline
    Real & 72.0 & 66.8 \\
    %\hline
    Sim & 67.8 & 66.1 \\
    %\hline
\bottomrule[0.1em]
    \end{tabular}
    \end{center}
    \caption{Detector trained on real data and evaluated on simulated vs. real data.}
    \label{tab::sim_eval}
    
    \begin{center}
    \begin{tabular}{ccc}
\toprule[0.1em]    
%\hline
    Train data (100k) & $\text{AP}_{0.3}$ & $\text{AP}_{0.5}$  \\
\midrule    
%\hline
    %\hline
    Real & 72.0 & 66.8 \\
    %\hline
%    Sim & 67.4 & 61.9 \\
%    \hline
    Real + Sim & 73.6 & 68.5 \\
    %\hline
\bottomrule[0.1em]
    \end{tabular}
    \end{center}
    \caption{Trained on 100k real and 100k sim (same layout) and evaluated on real data.}
    \label{tab::sim_train}
\end{minipage}
\hfill
\begin{minipage}{0.55\textwidth}
    \begin{center}
    \begin{tabular}{cccccc}
\toprule[0.1em]    
%\hline
    \multirow{2}{*}{Real Amount} & \multicolumn{2}{c}{~~Real Only~~} & & \multicolumn{2}{c}{Real+100k Sim} \\
    \cmidrule{2-3}\cmidrule{5-6}
%\cline{2-3}\cline{5-6}
    & $\text{AP}_{0.3}$ & $\text{AP}_{0.5}$ & & $\text{AP}_{0.3}$ & $\text{AP}_{0.5}$ \\
\midrule   
% \hline
    %\hline 
    0k & -- & -- && 66.9 & 61.6 \\
    %\hline
    5k & 30.9 & 27.5 && 68.7 & 63.2 \\
   % \hline
    10k & 40.2 & 36.6 && 69.4 & 64.3 \\
    %\hline
    20k & 53.2 & 48.6 && 70.4 & 65.4\\
    %\hline
    50k & 67.4 & 62.7 && 73.4 & 68.5 \\
    %\hline
    100k & 72.0 & 66.8 && 74.9 & 69.9 \\
    %\hline
\bottomrule[0.1em]
    \end{tabular}
    \end{center}
    \caption{Training with simulated data boosts pedestrian detection performance.}
    \label{tab::sim_ablation}
\end{minipage}
\vspace{-20pt}
\end{table}

\paragraph{Evaluating Pedestrian Detector on Simulated Data:}
We first evaluate the pedestrian detector on our simulated LiDAR data, and compare the result with the one evaluated on real LiDAR data.
This indicates how well we can use our simulation for testing the perception system on safety-critical scenarios.
To properly evaluate the realism of our simulated LiDAR points, we generate LiDARsim point clouds from the ground-truth scene layouts.
The pedestrian detector was trained on real LiDAR data only.
We evaluate the average precision (AP) of the detector for the pedestrian class at two IoU thresholds: $0.3$ and $0.5$.
As seen in Table~\ref{tab::sim_eval}, our method has a small gap of 0.7 points at IoU 0.5. This means we can directly use it with little to no domain adaptation to evaluate autonomy systems.

\vspace{-5pt}
\paragraph{Training Data Augmentation with Simulated Data:}
We train the detector on varying amounts of real LiDAR data and show how performance changes when we augment the dataset with 100k examples of simulated LiDAR data containing vehicles and pedestrians. 
We report the results in Table~\ref{tab::sim_ablation}. 
Note that to strictly evaluate the realism of the sensor data, the pedestrian layout and trajectory in the 100k simulated examples are different from that in the 100k real examples. 
When we combine simulated LiDAR data with real data, we consistently get performance gains, especially when we only have limited real data. Moreover, when we combine both large amounts of real and simulation data (100k examples each), we get about $3$ AP point improvement over real data alone.  
As shown in Table~\ref{tab::sim_train}, if 100k real LiDAR examples and 100k simulated LiDAR examples use the same scene layout and pedestrian trajectory, we get $1.7$ AP point improvement over real data alone, highlighting the value of simulating diverse pedestrian LiDAR sequences even with the same layout.

\vspace{-5pt}
\section{Conclusion}
\label{sec::conclusion}
\vspace{-10pt}

In this paper, we propose to leverage LiDAR and camera images collected by self-driving cars driving around a city to generate diverse pedestrian shapes and motions at scale, which we then use for accurate simulation to test and train a state-of-the-art perception system. 
Towards this goal, we have designed a deep-structured model, LiME, to reconstruct pedestrians in the wild using image and LiDAR sequences.
We then perform motion retargeting and pedestrian scenario simulation in urban scenes to generate realistic LiDAR data.
Our results show that the generated LiDAR point clouds have little domain gap and enhance the performance of downstream detectors via data augmentation. 
In the future we plan to reconstruct and simulate other categories such as animals.

%===============================================================================

\clearpage
% The acknowledgments are automatically included only in the final version of the paper.
%\acknowledgments{If a paper is accepted, the final camera-ready version will (and probably should) include acknowledgments. All acknowledgments go at the end of the paper, including thanks to reviewers who gave useful comments, to colleagues who contributed to the ideas, and to funding agencies and corporate sponsors that provided financial support.}

%===============================================================================

% no \bibliographystyle is required, since the corl style is automatically used.

% appendix
\clearpage
% !TEX root =  ../main.tex
\section*{Appendix}
\appendix
\renewcommand{\thesection}{A\arabic{section}}  
%===============================================================================
In this supplementary, we cover additional details and analysis of our method for recovering and simulating pedestrians in the wild.
In Section~\ref{sec:raytracer} we provide details about our obstacle-aware ray-tracer that allows us to incorporate LiDAR observations to improve 3D pose and shape reconstruction, and we discuss the differentiability of our ray-tracer.
Then in Section \ref{sec:learn} we provide more details about our learning and inference pipelines.
Finally in Section \ref{sec:detector} we provide the details of our pedestrian detector used in the experiments.

Additionally, please see our supplementary video, which showcases 
(1) Motivation and our methodology overview of LiME (LiDAR for human Mesh Estimation); 
(2) Human shape and pose reconstruction results using LiME on our real-world data, demonstrating the diversity of pedestrians we recover;
(3) Application of our pedestrian asset bank for downstream evaluation of perception algorithms trained only on real data; and
(4) Demonstration of our method for training perception algorithms by showing a side-by-side comparison of a detector trained on either simulated or real data and evaluated on real data.

\section{Details of our Obstacle-aware Differentiable Ray-tracer}
\label{sec:raytracer}

As described in the main paper, real LiDAR point cloud observations of pedestrians in the wild are sparse (due to distance and LiDAR resolution) and partial (due to occlusions of other objects).
The LiDAR sensor can be approximated via ray casting, where each laser ray shot by the sensor is parameterized in spherical coordinates $(r, \phi, \theta)$, representing the radius (distance travelled), azimuth, and elevation of the ray.  
We therefore design our ray-tracer to generate synthetic point clouds that better match the real ones by using the same LiDAR resolution when generating the ray-casted rays and removing ray-casted rays that hit occluded objects, which we can infer based on the real LiDAR point cloud.

\paragraph{Ray-casting algorithm:}
Given the pedestrian LiDAR point cloud and LiDAR sensor location, we first compute the radius, azimuth, and elevation ranges of the rays that might hit the pedestrian as $\{r_\text{min}, r_\text{max}\}$, $\{\phi_\text{min}, \phi_\text{max}\}$ and $\{\theta_\text{min}, \theta_\text{max}\}$.
We determine these values based on the 3D bounding box enclosing the pedestrian LiDAR point clud.
We then compute the set of rays within the azimuth and elevation range according to the resolution of LiDAR sensor $(d_\phi, d_\theta)$:
\begin{align}
	\mathcal R = \left\{ 
	\left( i*d_\phi,  j*d_\theta \right) \middle\vert\
	\lfloor \frac{\phi_\text{min}}{d_\phi} \rfloor<i<\lfloor \frac{\phi_\text{max}}{d_\phi} \rfloor,
	\lfloor \frac{\theta_\text{min}}{d_\theta} \rfloor<j<\lfloor \frac{\phi_\text{max}}{d_\phi} \rfloor
	\right\}
\end{align}
where $\lfloor \cdot \rfloor$ is the floor function, and $i,j$ are integers. 
For each ray $\mathbf r=(i*d_\phi,  j*d_\theta) \in \mathcal R$, $i*d_\phi$ is the azimuth of the ray and $j*d_\theta$ is the elevation of the ray. 
For simplicity, we assume the centre of the ray-caster is at origin $\mathbf{o}$.

We then cast the set of rays $\mathcal R$ into the reconstructed mesh using the Moller-Trumbore~\cite{RayCasting} ray casting algorithm. 
Moller-Trumbore ray casting efficiently computes the ray-triangle intersection for each triangle in the mesh by converting the representation of the intersection point $\mathbf{p}$ from cartesian coordinates to the Barycentric coordinates of the triangle of interest.
We define the cartesian coordinate of the intersection point as $\mathbf{p_{cart}} = \mathbf{o} + c~\mathbf{d}$, where $\mathbf{o}$ and $\mathbf{d}$ are the origin and direction of the raycasted ray in $(x,y,z)$ cartesian coordinate space, and $c$ is the distance travelled.
For a triangle face $\mathbf f$ with vertices $(\mathbf v_1, \mathbf v_2, \mathbf v_3)$, we define $\mathbf e_1 = \mathbf v_2 - \mathbf v_1, \mathbf e_2 = \mathbf v_3 - \mathbf v_2$ and $\mathbf t = \mathbf o - \mathbf v_1$. 
The Barycentric coordinates of the intersection point $(u, v)$ are obtained by solving: 
\begin{align}
   \label{eq::MT}
    \begin{bmatrix}
    c\\u\\v
    \end{bmatrix}
    = \frac{1}{(\mathbf d \times \mathbf e_2)\cdot \mathbf e_1} 
    \begin{bmatrix}
    (\mathbf t \times \mathbf e_1)\cdot \mathbf e_2\\
    (\mathbf d \times \mathbf e_2)\cdot \mathbf t\\
    (\mathbf t \times \mathbf e_1)\cdot \mathbf d
    \end{bmatrix}
\end{align}
where $\times$ is the cross product operator and $\cdot$ is the inner product operator between two vectors. If the intersection point lies inside the triangle, we can convert the intersection point back to cartesian coordinates as $\mathbf y = \mathbf v_1 + u \mathbf e_1 + v \mathbf e_2$. 
Note that if the ray intersect with multiple triangle faces, we choose the ray-casted point with \textbf{minimum} distance to the ray-caster origin.
The ray-casted points on the mesh form the set $\mathbf Y$ (Eq.~\ref{eq::sim} in the main paper).

\paragraph{Occlusion-aware ray-caster:} 

Directly using a ray-tracer to generate a synthethic point cloud will not match well with the observed LiDAR points if the set of rays $\mathcal R$ hit occluded objects in the real LiDAR scene. 
Not accounting for these occlusions will incorrectly penalize the posed mesh to not have points generated in these regions not visible to the real LiDAR sensor.
To account for occlusions, we first
define 
an occluded object
as an object in front of sensor path to the bounding box enclosing the pedestrian LiDAR points. 
Note that we do not account for occlusion inside the bounding box. 
Then the set of points in the real LiDAR scan forming the occlusion is:
\begin{align}
\mathbf O = \{ \mathbf p ~|~ r_{\mathbf p} <r_\text{min}, \phi_\text{min} < \phi_{\mathbf p} <\phi_\text{max}, \theta_\text{min} < \theta_{\mathbf p} <\theta_\text{max}\}	
\end{align}
where $r_{\mathbf p}, \phi_{\mathbf p}, \theta_{\mathbf p}$ are the radius, azimuth and elevation of point $\mathbf p$, respectively. We determine the rays in $\mathcal R$ that hit the occlusion $\mathbf O$ as:
\begin{align}
	\mathcal O = \left\{ 
	\left( 
	\lfloor \frac{\phi_{\mathbf p}}{d_\phi} \rfloor,
	\lfloor \frac{\theta_{\mathbf p}}{d_\theta} \rfloor
	\right) \middle\vert\
	\mathbf p \in \mathbf O
	\right\}
\end{align}

\begin{figure}[t]
	\centering
    \includegraphics[width = 0.6\linewidth]{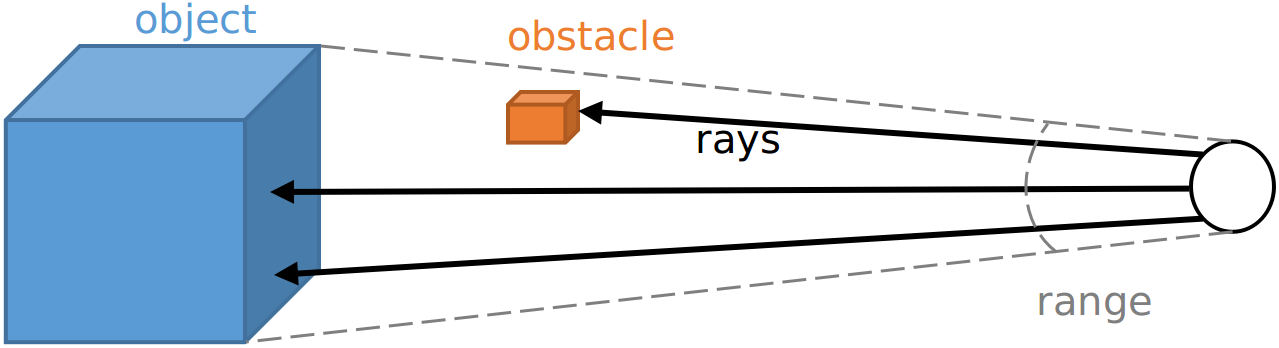}
    \caption{
    We determine the rays to be casted using the bounding box enclosing the object LiDAR points (\textcolor{blue}{blue}), and we mask out the rays that hit obstacles (\textcolor{orange}{orange}). We use the remaining rays to compute the ray-casted points.
	}
    \label{fig:vis_raytracer}
\end{figure}

We then mask out occluded rays $\mathcal O$ from $\mathcal R$ and use the set of rays $\mathcal R \setminus \mathcal O$ to generate the ray-casted points $\mathbf Y$, and compute $E_{\text{sim}}$ (Eq.~\ref{eq::sim} in the main paper):
\begin{align}
    \label{eq::sim_extend}
    E_{\text{sim}}(\mathbf \Theta_t, \mathbf c_t, \mathbf s,  \mathbf D) &= \frac{1}{|\mathbf X|} \sum_{\mathbf x \in \mathbf X} \min_{\mathbf y \in \mathbf Y} \left\lVert \mathbf x - \mathbf y \right\rVert_2^2 +
\frac{1}{|\mathbf Y|} \sum_{\mathbf y \in \mathbf Y} \min_{\mathbf x \in \mathbf X} \left\lVert \mathbf y - \mathbf x \right\rVert_2^2
\end{align} 
See Figure~\ref{fig:vis_raytracer} for a visual explanation.

\paragraph{Differentiability of the ray-tracer.}
Since the coordinate of the ray-casted point is:
\begin{align}
  	\nonumber \mathbf y &= \mathbf v_1 + u \mathbf e_1 + v \mathbf e_2\\
  	\nonumber &=\mathbf v_1 + u (\mathbf v_2 - \mathbf v_1) + v (\mathbf v_3 - \mathbf v_2)\\
  	&= (1-u) \mathbf v_1 + (u-v) \mathbf v_2 + v \mathbf v_3
\end{align}
which is a linear combination of the vertices $\mathbf v_1, \mathbf v_2, \mathbf v_3$ in the mesh. The LiDAR ray-tracer is differentiable with respect to the mesh vertices. 
Although this ray-caster is not differentiable with respect to which triangle it intersects, empirically we find this works well in practice, as we have additional energy terms such as the 2D-joints consistency from the image to provide key-points correspondence supervision, and shape and pose priors to help with the shape and pose estimates.

\section{Learning and Inference Details}
\label{sec:learn}
In our Sensor Fusion Network, we use ResNet-50 as the 2D CNN backbone, and we use the Point Completion Network~\cite{PCN} as the Point Cloud feature extractor. To learn the neural network, we use batch size of 16, Adam optimizer with learning rate $1e-4$. And we train the network for $50000$ iterations.

When we perform energy minimization, we use the Adam optimizer with learning rate of $1e-2$.
The weight for simulation, joints, pose prior, bone scale prior, L2 smoothness and Laplacian term are $144^2$, $0.2^2$, $0.478^2$, $2^2$, $100^2$, $1000^2$, respectively.

\section{Pedestrian Detector Details}
\label{sec:detector}
We use the object detector similar to~\cite{pnpnet}, it takes as input five consecutive LiDAR sweeps (0.5s) in birds-eye-view (BEV). The LiDAR data uses a voxel based representation in BEV, and the five consecutive sweeps are combined by concatenating along the height dimension (with the ego motion compensated for the previous sweeps). Each instance label box includes at least one pedestrian LiDAR point.

Given the aforementioned BEV representation of the LiDAR as input, the network first down-sample the input BEV image by a factor of 4 using three Conv2D layers. Then a cross-scale module~\cite{pnpnet} was applied three times sequentially. Next, a FPN was applied to fuse multi-scale feature maps, resulting in a 4x down-sampled BEV feature map. Finally we use 4 Conv2D layers to generate 3D bounding box prediction.

\end{document}